\documentclass[11pt]{article}
\usepackage{coling2018}
\usepackage{times}
\usepackage{url}
\usepackage{latexsym}

\usepackage{amsmath}
\usepackage{amsthm}
\usepackage{amssymb}
\usepackage{hyperref}
\usepackage{colortbl}
\usepackage{graphicx}
\usepackage{xcolor}
\usepackage{multirow}
\usepackage{todonotes}

\newcommand{\comment}[1]{}

\newcommand{\pce}{\mbox{pce}}

\newcommand{\KL}{\mbox{KL}}

\title{Self-Normalization Properties of Language Modeling }

\author{Jacob Goldberger \\
  Faculty of Engineering \\
Bar-Ilan University, Israel  \\
  {\small \tt jacob.goldberger@biu.ac.il} \\\And
  Oren Melamud \\
  IBM Research  \\
  Yorktown Heights, NY, USA  \\
  {\small \tt oren.melamud@ibm.com} \\}

\date{}

\begin{document}

\maketitle
\begin{abstract}
\emph{Self-normalizing} discriminative  models approximate the normalized probability of a class without having to compute the partition function.
In the context of language modeling, this property 
is particularly appealing as it may significantly reduce run-times due to large word vocabularies.
In this study, we provide a comprehensive investigation of language modeling self-normalization. First, we theoretically analyze the inherent self-normalization properties of Noise Contrastive Estimation (NCE) language models. Then, we compare them empirically to softmax-based approaches, which are self-normalized using explicit regularization, and suggest a hybrid model with compelling properties. Finally, we uncover a surprising negative correlation between self-normalization and perplexity across the board, as well as some regularity in the observed errors, which may potentially be used for improving self-normalization algorithms in the future.

\end{abstract}

\section{Introduction}

%
% The following footnote without marker is needed for the camera-ready
% version of the paper.
% Comment out the instructions (first text) and uncomment the 8 lines
% under "final paper" for your variant of English.
% 
\blfootnote{
    %
    % for review submission
    %
%     \hspace{-0.65cm}  % space normally used by the marker
%     Place licence statement here for the camera-ready version. See
%     Section~\ref{licence} of the instructions for preparing a
%     manuscript.
    %
    % % final paper: en-uk version 
    %
    % \hspace{-0.65cm}  % space normally used by the marker
    % This work is licenced under a Creative Commons 
    % Attribution 4.0 International Licence.
    % Licence details:
    % \url{http://creativecommons.org/licenses/by/4.0/}
    % 
    % % final paper: en-us version 
    %
    \hspace{-0.65cm}  % space normally used by the marker
    This work is licensed under a Creative Commons 
    Attribution 4.0 International License.
    License details:
    \url{http://creativecommons.org/licenses/by/4.0/}
}

The ability of statistical language models (LMs) to estimate the probability
of a word given a context of preceding
words, plays an important role in many NLP
tasks, such as speech recognition and machine translation. Recurrent Neural Network (RNN) language models have recently become the preferred method of choice, having outperformed traditional $n$-gram LMs across a range of tasks \cite{jozefowicz2016exploring}.
Unfortunately however, they suffer from scalability issues incurred by the computation of the softmax normalization term, which is required to guarantee proper probability predictions. The cost of this computation is linearly proportional to the
size of the word vocabulary and has a significant
impact on both training and testing run-times.

Several methods have been proposed to cope with this scaling issue by replacing the softmax with a more computationally efficient component at train time. \footnote{Alleviating this problem using sub-word representations is a parallel line of research not discussed here.}
These include importance sampling \cite{Bengio2003}, hierarchical softmax \cite{Mnihnips}, BlackOut \cite{ji2016blackout} and Noise Contrastive Estimation (NCE) \cite{Gutmann2012}.
NCE has been applied to train neural LMs with large vocabularies \cite{Mnih2012} and more recently was also successfully used to train LSTM-RNN LMs \cite{Vaswani2013,Chen2015,Zoph2016}, achieving near state-of-the-art performance on language modeling tasks \cite{jozefowicz2016exploring,Chen2016StrategiesFT}.
All the above works focused on reducing the complexity at train time.
However, at test time, the assumption was that one still needs to compute the costly softmax normalization term to obtain a normalized score fit as an estimate for the probability of a word.

\emph{Self-normalization} was recently proposed to address the test time complexity. A self-normalized discriminative model is trained to produce near-normalized scores in the sense that the sum over the scores of all words is approximately  one. If this approximation is close enough, the assumption is that the costly exact normalization can be waived at test time without significantly sacrificing prediction accuracy \cite{Devlin}.
Two main approaches were proposed to train self-normalizing models. Regularized softmax self-normalization is based on using softmax for training and explicitly
encouraging the normalization term of the softmax to be as close to one as possible, thus making its computation redundant at test time
\cite{Devlin,Andreas_2015,Chen2016StrategiesFT}. The alternative approach is based on NCE.
The original formulation of NCE included a parametrized normalization term $Z_c$ for every context $c$. However, the first work that applied NCE to language modeling  \cite{Mnih2012} discovered empirically that fixing $Z_c$ to a constant did not affect the performance. More recent studies
\cite{Vaswani2013,Zoph2016,Chen2015,Oualil} empirically found that models trained using NCE with a fixed $Z_c$, exhibit self-normalization at test time. This behavior is facilitated by inherent self-normalization properties of NCE LMs that we analyze in this work.

The main contribution of this study is in providing a first comprehensive investigation of self-normalizing language models. This includes a theoretical analysis of the inherent self-normalizing properties of NCE language models, followed by an extensive empirical evaluation of NCE against softmax-based self-normalizing methods.
Our results suggest that regularized softmax models perform competitively as long as we are only interested in low test time complexity. However, when train time is also a factor, 
NCE has a notable advantage.
Furthermore, we find, somewhat surprisingly, that larger models that achieve better perplexities tend to have worse self-normalization properties, and perform further analysis in an attempt to better understand this behavior.
Finally, we show that downstream tasks may not all be as sensitive to self-normalization as might be expected.

The rest of this paper is organized as follows. In sections \ref{sec:section1} and \ref{sec:selfnormprop}, we provide theoretical background and analysis of NCE language modeling that justifies its inherent self-normalizing properties. In Section~\ref{sec:explicit}, we review the alternative regularized softmax-based self-normalization methods and introduce a novel regularized NCE hybrid approach.
In Section \ref{sec:intrinsic}, we report on an empirical intrinsic investigation of all the methods above, and finally, in Section \ref{sec:mscc}, we evaluate the compared methods on the  Microsoft's Sentence Completion Challenge and compare these results with the intrinsic measures of perplexity and self-normalization.

\section{NCE as a Matrix Factorization}
\label{sec:section1}
In this section, we review the NCE algorithm for language modeling  \cite{Gutmann2012,Mnih2012} and focus on its interpretation as a matrix factorization procedure. This analysis is analogous to the one proposed by \newcite{Melamud_emnlp} for their PMI language model.
Let $p(w|c)$ be the probability of a word $w$ given a preceding context $c$, and let $p(w)$ be the  word unigram distribution. Assume the distribution $p(w|c)$
has the following parametric form:
\begin{equation}
p_{nce}(w|c) =
\frac{1}{Z_c} \exp( m(w,c) )
\label{loglinearm}
\end{equation}
such that $m(w,c) = \vec{w} \cdot \vec {c}+b_w$, where   $\vec{w}$ and $\vec{c}$ are $d$-dimensional vector representations of  the word $w$  and its context $c$, and $Z_c$ is a normalization term.

We can use a
simple lookup table for the word representation $\vec{w}$,
and a recurrent neural network (RNN) model to obtain a
low dimensional representation of the entire preceding
context $\vec{c}$.
Given a text corpus~$D$, the NCE objective function is:
\begin{equation} S(m)=\sum_{w,c \in D} \Big[  \log \sigma ( m(w,c) -\log (p(w) k) )
\label{eq:nceobjun}
\end{equation}
$$+\sum_{i=1}^k \log  (1- \sigma( m(w_i,c) - \log (p(w_i) k)))\Big]$$
such that $w,c$ go over all the word-context co-occurrences in the learning corpus $D$ and $w_1,...,w_k$ are `noise' samples drawn from the word unigram distribution. $\sigma$ denotes the sigmoid function.

Let  $\pce(w,c) = \log p(w|c)$ be the Pointwise Conditional Entropy (PCE) matrix, which is the true log probability we are trying to estimate.  \newcite{Gutmann2012} proved that $S(m)  \le S(\pce)$ for every matrix $m$.
The rank of the matrix $m$ is at most $d+1$.
Thus, the NCE training goal is finding the best  low-rank decomposition of the PCE matrix in the sense that it minimizes the difference $S(\pce)-S(m)$. Following \newcite{melamud2017acl}, we can explicitly write this difference as a Kullback-Leibler~(KL) divergence.
The NCE derivation was originally based on sampling $w$ and $c$ either from the joint distribution or from the  product of marginals according to a  binary r.v. denoted by $z$. 
For every matrix~$m$, the conditional distribution of $z$ given  $w$ and $c$ is: 
 $$
p_m(z\!=\!1|w,c) = \sigma(m(w,c)-\log(kp(w))).
 $$
 The difference between the NCE score at the PCE matrix and the NCE score at a given matrix $m$ can be written as:
\begin{equation} S(\pce)- S(m) = \KL ( p_{\pce}(z|w,c) || p_m(z|w,c))
\label{factcrit}
\end{equation}
$$
=\sum_{w,c} p(w,c) \sum_{z=0,1} p_{\pce}(z|w,c) \log \frac{p_{\pce}(z|w,c)}{p_m(z|w,c)}.
$$

This view of NCE as a matrix factorization instead of a distribution estimation, makes the normalization factor redundant during training, thereby justifying the heuristics of setting $Z_c=1$ used by \newcite{Mnih2012}. The crux of the matrix decomposition view of NCE is that although the normalization term  is not explicitly included here, the optimal low-dimensional model attempts to approximate the true conditional probabilities, which are normalized, and therefore we expect that it will be almost self-normalized.
Indeed, in the next section we provide  formal guarantees for that.

\section {The NCE Self-Normalization property}
\label{sec:selfnormprop}
We now address the test time efficiency of language models, which is the focus of this study.
As is the case with other language models, at test time, when we use the low-dimensional matrix learned by NCE to compute the conditional probability  $p_{nce}(w|c)$ (\ref{loglinearm}),
  we need to compute the normalization factor to obtain a valid distribution:
\begin{equation}
Z_c = \sum_w \exp (m(w,c)) = \sum_w \exp (\vec{w} \cdot \vec{c} +b_w ).
\end{equation}
Unfortunately, this computation of $Z_c$ is often very expensive due to the typical large vocabulary size. However, as we next show, for NCE language models this computation may be avoided not only at train time, but also at test time due to self-normalization.

A matrix  $m$ is called self-normalized if
$\sum_w (\exp(m(w,c))=1$ for every~$c$.
The full-rank optimal  LM obtained from the PCE matrix $\pce(w,c) = \log p(w|c)$,  is clearly self normalized:
$$
Z_c = \sum_w \exp( \pce(w,c) ) = \sum_w    p(w|c)=1.
$$
 The NCE algorithm seeks the best low-rank unnormalized matrix approximation of the PCE matrix. Hence, we can assume that the matrix $m$ is close to the  PCE  matrix and therefore defines a LM that
should also be close to self-normalized:
\begin{equation}
\sum_w \exp( m(w,c) )  \approx \sum_w \exp( \pce(w,c))  = 1.
\label{self_pce}
\end{equation}

We next formally show that if the matrix  $m$ is close to the PCE matrix then the NCE model defined by $m$ is approximately self-normalized.

{\bf Theorem 1:} Assume that for a given context $c$ there is an $0<\epsilon $ such that $$\log \sum_{w \in V} p(w|c) \exp (  |m(w,c)- \log p(w|c)|) \le \epsilon.$$
Let $Z_c= \sum_w \exp(m(w,c))$ be the normalization factor. Then $|\log Z_c |   \le   \epsilon.$

{\bf Proof:}
$$ \log Z_c  =  \log \sum_w \exp ( \vec{w} \cdot \vec{c} +b_w )  $$ $$   =   \log \sum_w ( p (w|c)  \exp (m(w,c)- \log p(w|c) ))  $$
\begin{equation}
\le \log \sum_{w} p(w|c) \exp (  |m(w,c)- \log p(w|c)|) \le \epsilon.
\label{eqap}
\end{equation}
The concavity of the log function implies that:
\begin{equation}  -\log Z_c \le   -\sum_w p(w|c)   ( m(w,c) - \log p(w|c))
\label{eqan}
\end{equation}
$$
= \sum_w p(w|c)  ( -(m(w,c) - \log p(w|c)))
$$
The convexity of the exp function implies that:
$$
\le \log  \sum_w p(w|c) \exp ( -(m(w,c)- \log p(w|c)))
$$
$$
\le
\log  \sum_w p(w|c) \exp ( |m(w,c) - \log p(w|c)|) \le \epsilon
$$
Combining Eq. (\ref{eqap}) and Eq. (\ref{eqan}) we finally obtain that $|\log Z_c |   \le   \epsilon.$ \hspace{1cm} $\Box $

We can also state a global version of Theorem 1 and its proof is similar.

{\bf Theorem 2:} Assume there is an $0<\epsilon $ such that $$\log \sum_{w,c} p(w,c) \exp (  |m(w,c)- \log p(w|c)|) \le \epsilon.$$
Then $|\sum_c p(c) \log Z_c |   \le   \epsilon.$

\section{Explicit Self-normalization}
\label{sec:explicit}

In this section, we review the two recently proposed language modeling methods that achieve self-normalization via explicit regularization, and then borrow from them to derive an novel regularized version of NCE. 

The standard language modeling learning method, which is based on a softmax output layer, is not self-normalized. To encourage its self-normalization, \newcite{Devlin} proposed to add to its training objective function, an explicit penalty for deviating from self-normalization:
\begin{equation}
  S_{Dev}=\sum_{w,c \in D} \Big[  (\vec{w} \cdot \vec {c} +b_w-\log Z_c)  - \alpha (\log Z_c)^2 \Big]
  \label{Devlinobj}
 \end{equation}
where $Z_c=\sum_{v \in V} \exp (\vec{v} \cdot \vec {c}+b_v)$ and $\alpha$ is a constant.
The drawback of this approach is that at train time you still need to explicitly compute the costly $Z_c$.
\newcite{Andreas_2015} proposed a more computationally efficient approximation of~(\ref{Devlinobj}) that eliminates  $Z_c$ in the first term and computes the second term only on a sampled subset $D'$ of the corpus $D$:
\begin{equation}
  S_{And}=\sum_{w,c \in D}   (\vec{w} \cdot \vec {c} +b_w)  - \frac{\alpha}{\gamma} \sum_{c \in D'}    (\log Z_c)^2
  \label{Andobj}
 \end{equation}
where $\gamma < 1$ is an additional constant that determines the sampling rate, i.e. $|D'| = \gamma|D|$.
They also provided analysis that justifies computing $Z_c$ only on a subset of the corpus by showing  that if a given LM is exactly self-normalized on a dense set of contexts (i.e. each context $c$ is close to a context $c'$ s.t. $\log Z_{c'}=0$) then $E|\log  Z_c |$ is small.

Inspired by this work, we propose a regularized variant of the NCE objective function~(\ref{eq:nceobjun}):
\begin{equation} S_{nce-r}(m) =S_{nce}(m) - \frac{\alpha}{\gamma} \sum_{c \in D'}    (\log Z_c)^2
\label{eq:ncerobj}
\end{equation}
This formulation allows us to further encourage the NCE self-normalization, still without incurring the cost of computing $Z_c$ for every word in the learning corpus.

\section{Intrinsic Evaluation}
\label{sec:intrinsic}

We report here on an empirical investigation of the self-normalization properties of NCE language modeling as compared to the alternative methods described in the previous sections.

\subsection{Experimental Settings}
\label{subsec:experimental_settings}

We investigated the following language modeling methods:

\begin{itemize}

\item \emph{DEV-LM} - the language model proposed by \newcite{Devlin} (Eq. \ref{Devlinobj})

\item \emph{SM-LM} - a standard softmax language model (DEV-LM with $\alpha=0$)

\item \emph{AND-LM} - the light-weight approximation of DEV-LM proposed by \newcite{Andreas_2015} (Eq. \ref{Andobj})

\item \emph{NCE-LM} - NCE language model (Eq. \ref{eq:nceobjun})

\item \emph{NCE-R-LM} - our light-weight regularized NCE method (Eq. \ref{eq:ncerobj})

\end{itemize}

Following \newcite{Devlin}, to make all of the above methods approximately self-normalized at init time, we initialized their output bias terms to $b_{w} = -\log|V|$, where $V$ is the word vocabulary. We set the negative sampling parameter for the NCE-based LMs to $k=100$, following \newcite{Zoph2016}, who showed highly competitive performance with NCE LMs trained with this number of samples, and following \newcite{Melamud_emnlp} who used the same with PMI language models. We note that in early experiments with PMI LMs, which can be viewed as a close variant of NCE-LMs, we got very similar results for both of these types of models and therefore did not include PMI-LMs in our final investigation.

All of the compared methods use standard LSTM to represent the preceding (left-side) sequence of words as the context vector $\vec{c}$, and a simple word embedding lookup table to represent the predicted next word as $\vec{w}$.
The LSTM hyperparameters and training regimen are similar to \newcite{zaremba2014recurrent} who achieved strong perplexity results compared to other standard LSTM-based neural language models. Specifically, we used a 2-layer LSTM with a 50\% dropout ratio. 
During training, we performed truncated back-propagation-through-time,
unrolling the LSTM for 20 steps at a time without ever resetting the LSTM state. We
trained our model for 20 epochs using Stochastic Gradient Descent (SGD) with a learning rate of 1, which is decreased by a factor of 1.2 after every epoch starting after epoch 6. We clip the norms of the gradient to 5 and use mini-batch size of 20.
All models were implemented using the Chainer toolkit (\cite{tokuichainer}).

We used two popular language modeling datasets in the evaluation.
The first dataset, denoted \emph{PTB},  is a version of the Penn Tree Bank, commonly used to evaluate language models.\footnote{Available from Tomas Mikolov at: \url{http://www.fit.vutbr.cz/~imikolov/rnnlm/simple-examples.tgz}} It consists of 929K/73K/82K training/validation/test words respectively and has a 10K word vocabulary. The second dataset, denoted \emph{WIKI}, is the WikiText-2, more recently introduced by \newcite{merity2016pointer}. This dataset was extracted from Wikipedia articles and is somewhat larger, with 2,088K/217K/245K train/validation/test tokens, respectively, and a vocabulary size of 33K.

To evaluate self-normalization, we look at two metrics: (1) $\mu_z = E(\log(Z_c))$, which is the mean log value of the normalization term, across the contexts in the evaluated dataset; and (2) $\sigma_z = \sigma(\log(Z_c))$, which is the corresponding standard deviation. The closer these two metrics are to zero, the more self-normalizing the model is considered to be. We note that a model with an observed $|\mu_z| >> 0$ on a dev set, can be `corrected' to a large extent (as we show later) by subtracting this dev $\mu_z$ from the unnormalized scores at test time. However, this is not the case for $\sigma_z$. Therefore, from a practical point of view, we consider $\sigma_z$ to be the more important metric of the two. In addition, we also look at the classic perplexity metric, which is considered a standard intrinsic measure  for the quality of the model predictions. Importantly, when measuring perplexity, except where noted otherwise, we first perform exact normalization of the models' unnormalized scores by computing the normalization term.

\subsection{Results}

\begin{table*}[t]
\centering
\begin{tabular}{| r || c | c | r || c | c | r |}
\hline
  & \multicolumn{3}{| c ||}{\emph{NCE-LM}} &  \multicolumn{3}{| c |}{\emph{SM-LM}} \\
\hline
 d   & $\mu_z$ & $\sigma_z$ & perp & $\mu_z$ & $\sigma_z$ & perp  \\
\hline
\rowcolor[gray]{.9}
 &  \multicolumn{6}{| c |}{PTB validation set} \\

\hline
30 & -0.18 & 0.11	& 267.6	& 2.29 & 	0.97	& 243.4 \\
100 & -0.19	& 0.17	& 150.9 &	3.03 &	1.52 &	145.2 \\
300 & -0.15	& 0.29 & 	100.1	& 3.77	& 1.98 &	97.7 \\
650 & -0.17 & 	0.37 &	87.4	& 4.36	& 2.31 &	87.3 \\
\hline
    \rowcolor[gray]{.9}
 &  \multicolumn{6}{| c |}{WIKI validation set} \\
\hline
30 & -0.20	& 0.13	& 357.4	& 2.57	& 1.02	& 322.2 \\
100 & -0.24	& 0.19	& 194.3	& 3.34	& 1.45	& 191.1 \\
300 & -0.23	& 0.27	& 125.6	& 4.19	& 1.73	& 123.3 \\
650 & -0.23	& 0.35	& 110.5	& 4.67	& 1.83	& 110.7 \\

 \hline
\end{tabular}
\caption{Self-normalization and perplexity results of NCE-LM against the standard softmax language model, SM-LM. $d$ denotes the size of the compared models (units).
}
\label{tab:nce_softmax_results}
\end{table*}

We begin by comparing the results obtained by the two methods that do not include any explicit self-normalization component in their objectives, namely NCE-LM and the standard softmax SM-LM.
Table~\ref{tab:nce_softmax_results} shows
that consistent with previous works, NCE-LM is approximately self-normalized as apparent by relatively low $|\mu_z|$ and $\sigma_z$ values. On the other hand, SM-LM, as expected, is much less self-normalized. In terms of perplexity, we see that SM-LM performs a little better than NCE-LM when model dimensionality is low, but the gap closes entirely at $d=650$. Curiously, while perplexity improves with higher dimensionality, we see that the quality of NCE-LM's self-normalization, as evident particularly by $\sigma_z$, actually degrades. This is surprising, as we would expect that stronger models with more parameters would approximate the true $p(w|c)$ more closely and hence be more self-normalized. A similar behavior was recorded for SM-LM.
We further investigate this in Section \ref{sec:analysis}.

We also measured model test run-times, running on a single Tesla K20 GPU. We compared run-times for normalized scores that were produced by applying exact normalization versus unnormalized scores. For both SM-LM and NCE-LM, which perform the same operations at test time, we get $\sim$9 seconds for normalized scores vs. $\sim$8 seconds for unnormalized ones on the PTB validation set. Run-times on the x3 larger Wiki validation set are $\sim$38 seconds for normalized and $\sim$24 seconds for unnormalized. We see that the run-time of the unnormalized models seems to scale linearly with the size of the dataset. However, the normalized run-time scales super-linearly, arguably since it depends heavily on the vocabulary size, which is greater for Wiki than for PTB. With typical vocabulary sizes reaching much higher than Wiki's 33K word types, this reinforces the computational motivation for self-normalized language models.

\begin{table*}[t]
\centering
\begin{tabular}{| r || c | c | r || c | c | r || c | c | r |}
\hline
  & \multicolumn{9}{| c |}{\emph{DEV-LM}} \\
\hline
\rowcolor[gray]{.9}
  & \multicolumn{3}{| c ||}{$\alpha = 0.1$} &  \multicolumn{3}{| c ||}{$\alpha = 1.0$} &  \multicolumn{3}{| c |}{$\alpha = 10.0$} \\
\hline
 d   & $\mu_z$ & $\sigma_z$ & perp & $\mu_Z$ & $\sigma_z$ & perp  & $\mu_z$ & $\sigma_z$ & perp \\
\hline
\rowcolor[gray]{.9}
 &  \multicolumn{9}{| c |}{PTB validation set} \\

\hline

30 & -0.12	& 0.21	& 242.6	& -0.16	& 0.09	& 250.9	& -0.13	& 0.060	& 307.2 \\
100 & -0.10	& 0.28	& 143.3	& -0.17	& 0.11	& 149.5	& -0.12	& 0.058	& 182.0 \\
300 & -0.09	& 0.36	& 96.3	& -0.14	& 0.14	& 100.8	& -0.16	& 0.054	& 121.3 \\
650 & -0.14	& 0.43	& 85.0	& \textbf{-0.17}	& \textbf{0.18}	& \textbf{86.3}	& -0.11	& 0.071	& 99.5 \\
\hline
    \rowcolor[gray]{.9}
 &  \multicolumn{9}{| c |}{WIKI validation set} \\
\hline

30 & -0.10	& 0.23	& 334.1	& -0.17	& 0.08	& 338.7	& -0.15	& 0.055	& 389.0 \\
100 & -0.13	& 0.28	& 189.4	& -0.22	& 0.13	& 191.1	& -0.15	& 0.071	& 228.3 \\
300 & -0.15	& 0.34	& 121.9	& -0.20	& 0.17	& 125.7	& -0.13	& 0.081	& 143.6 \\
650 & -0.23	& 0.42	& 109.1	& \textbf{-0.23}	& \textbf{0.20}	& \textbf{110.0}	& -0.12	& 0.089	& 116.9 \\

 \hline
\end{tabular}
\caption{Self-normalization and perplexity results of the self-normalizing DEV-LM for different values of the normalization factor $\alpha$. $d$ denotes the size of the compared models (units).
}
\label{tab:devlin_results}
%\vspace{10 pt}
\end{table*}

Next, Table \ref{tab:devlin_results} compares the self-normalization and perplexity performance of DEV-LM for varied values of the constant $\alpha$ on the validation sets. As could be expected, the larger the value of $\alpha$ is, the better the self-normalization becomes, reaching very good self-normalization for $\alpha~=~10.0$. On the other hand, the improvement in self-normalization seems to occur at the expense of perplexity. This is particularly true for the smaller models,
but is still evident even for $d=650$. Interestingly, as with NCE-LM, we see that $\sigma_z$ grows (i.e. self-normalization becomes worse) with the size of the model, and is negatively correlated with the improvement in perplexity.

\begin{table*}[t]
\centering
\begin{tabular}{| r || c | c | r || c | c | r |}
\hline
  & \multicolumn{3}{| c ||}{NCE-R-LM} &  \multicolumn{3}{| c |}{AND-LM} \\
\hline
 $\alpha$   & $\mu_z$ & $\sigma_z$ & perp & $\mu_z$ & $\sigma_z$ & perp  \\
\hline
\rowcolor[gray]{.9}
 &  \multicolumn{6}{| c |}{PTB validation set} \\

\hline
0.1 & -0.19 & 0.34 & 87.1 & 6.14 & 0.56 & 117.5 \\
1.0 & -0.21 & 0.27 & 87.2 & \textbf{0.45} & \textbf{0.25} & \textbf{119.4} \\
10.0 & \textbf{-0.19} & \textbf{0.17} & \textbf{89.8} & -0.037 & 0.079 & 143.7 \\
100.0 & -0.089 & 0.086 & 112.6 & -0.024 & 0.030 & 209.5\\

\hline
    \rowcolor[gray]{.9}
 &  \multicolumn{6}{| c |}{WIKI validation set} \\
\hline

0.1 & -0.23 & 0.33 & 111.1 & \textbf{4.85} & \textbf{0.72} & \textbf{201.5} \\
1.0 & -0.24 & 0.28 & 107.5 & 1.02 & 0.001 & 1481.3\\
10.0 & \textbf{-0.22} & \textbf{0.19} & \textbf{110.8} & 0.41 & 0.12 & 33323.1 \\
100.0 & -0.12 & 0.099 & 131.5 & 0.413 &  0.000 & 33278.0\\

 \hline
\end{tabular}
\caption{Self-normalization and perplexity results of the self-normalizing DEV-LM for different values of the normalization factor $\alpha$. $d$ = 650 and $\gamma$ = 0.1.
}
\label{tab:andreas_results}
\end{table*}

Finally, in Table \ref{tab:andreas_results}, we compare AND-LM against our proposed NCE-R-LM, using a sampling rate of $\gamma = 0.1$ to avoid computing $Z_c$ most of the time, and varied values of $\alpha$. As can be seen, AND-LM exhibits relatively bad performance. In particular, to make the model converge when trained on the WIKI dataset, we had to follow the  heuristic suggested by \newcite{Chen2016StrategiesFT}, applying the following conversion to all of AND-LM's unnormalized scores, $x \rightarrow 10 \tanh(x/5)$. In contrast, we see that NCE-R-LM is able to use the explicit regularization to improve self-normalization at the cost of a relatively small degradation in perplexity.

\begin{table*}[t]
\centering
\begin{tabular}{| l || c | c |r |r || c | c | r | r |}
\hline
  & \multicolumn{4}{| c ||}{\emph{PTB-test}} &  \multicolumn{4}{| c |}{\emph{WIKI-test} } \\
\hline
   \rowcolor[gray]{.9}
    & $\mu_z$ & $\sigma_z$ & perp & u-perp & $\mu_z$ & $\sigma_z$ & perp  & u-perp\\
\hline
DEV-LM &	-0.001 &	0.17 &	83.1 & 83.0 &	0.002 &	0.20 &	104.1  &  104.2 \\
NCE-R-LM & 0.002 & 0.17 & 85.9 & 86.0 & -0.003 & 0.19 & 105.0 & 104.7\\
NCE-LM &	-0.004 &	0.35 &	83.7 &	83.4 &	0.003 &	0.36 &	104.3 &	104.6 \\
AND-LM & 0.001 & 0.30 & 114.9 & 115.0 & 0.018 & 0.74 & 185.7 & 189.1 \\

 \hline
\end{tabular}
\caption{Self-normalization and perplexity results on test sets for `shifted' models with $d=650$. `u-perp' denotes unnormalized perplexity.
}
\label{tab:test_results}
\end{table*}

Switching to the test-set evaluation, we propose a simple technique to center the $\log(Z)$ values of a self-normalizing model's scores around zero. Let $\mu_{z}^{valid}$ be $E(\log(Z))$ observed on the validation set at train time. The probability estimates of the `shifted' model at test time are $\log p(w|c) = \vec{w} \cdot \vec{c} +b_w - \mu_{z}^{valid}$. Table \ref{tab:test_results} shows the results that we get when evaluating the shifted versions of DEV-LM, NCE-R-LM, NCE-LM and AND-LM with $d~=~650$.
For each compared model, we chose the $\alpha$ value that showed the best self-normalization performance without sacrificing significant perplexity performance. Specifically, we used $\alpha_{\textsc{DEV-LM}}=1.0$ and $\alpha_{\textsc{NCE-R-LM}}=10.0$ for both PTB and WIKI datasets, and then $\alpha_{\textsc{AND-LM}}=1.0$ and $\alpha_{\textsc{AND-LM}}=0.1$ for the PTB and WIKI datasets, respectively.
Following \newcite{Oualil}, in addition to perplexity, we also report `unnormalized perplexity', which is computed with the unnormalized model scores. When the unnormalized perplexity measure is close to the real perplexity, this suggests that the unnormalized scores are in fact nearly normalized.

As can be seen, with the shifting method, all models achieve near perfect (zero) $\mu_z$ value, and their unnormalized perplexities are almost identical to their respective real perplexities. Also, with the exception of AND-LM, the perplexities of all models are nearly identical. Finally, the standard deviation of the normalization term of DEV-LM and NCE-R-LM is notably better than that of NCE-LM and AND-LM. DEV-LM and NCE-R-LM perform very similar in all respects. However, we note that NCE-R-LM's advantage is that during training, it performs sparse computations of the costly normalization term
and therefore its training time depends much less on the size of the vocabulary.

\begin{table*}[t]
\centering
\begin{tabular}{| r || c | c || c |c |}
\hline
  & \multicolumn{2}{| c ||}{\emph{PTB-validation}} &  \multicolumn{2}{| c |}{\emph{WIKI-validation} } \\
\hline
   \rowcolor[gray]{.9}
 d   & NCE-LM & DEV-LM ($\alpha=1.0$) & NCE-LM & DEV-LM ($\alpha=1.0$)  \\
\hline

30 & -0.33 & -0.27 & -0.50 & -0.26 \\
100 & -0.29 & -0.29 & -0.53 & -0.49 \\
300 & -0.46 & -0.41 & -0.56 & -0.63 \\
650 & -0.50 &  -0.45 & -0.53 & -0.64 \\

 \hline
\end{tabular}
\caption{Pearson's correlation between $H_c$ (entropy) and $\log(Z_c)$ on samples from the validation sets.
}
\label{tab:entropy_correlation}
\vspace{10 pt}
\end{table*}

\begin{figure*}[h!!!]
\centering

$
\begin{array}{ccc}
\includegraphics[width=5.0cm]{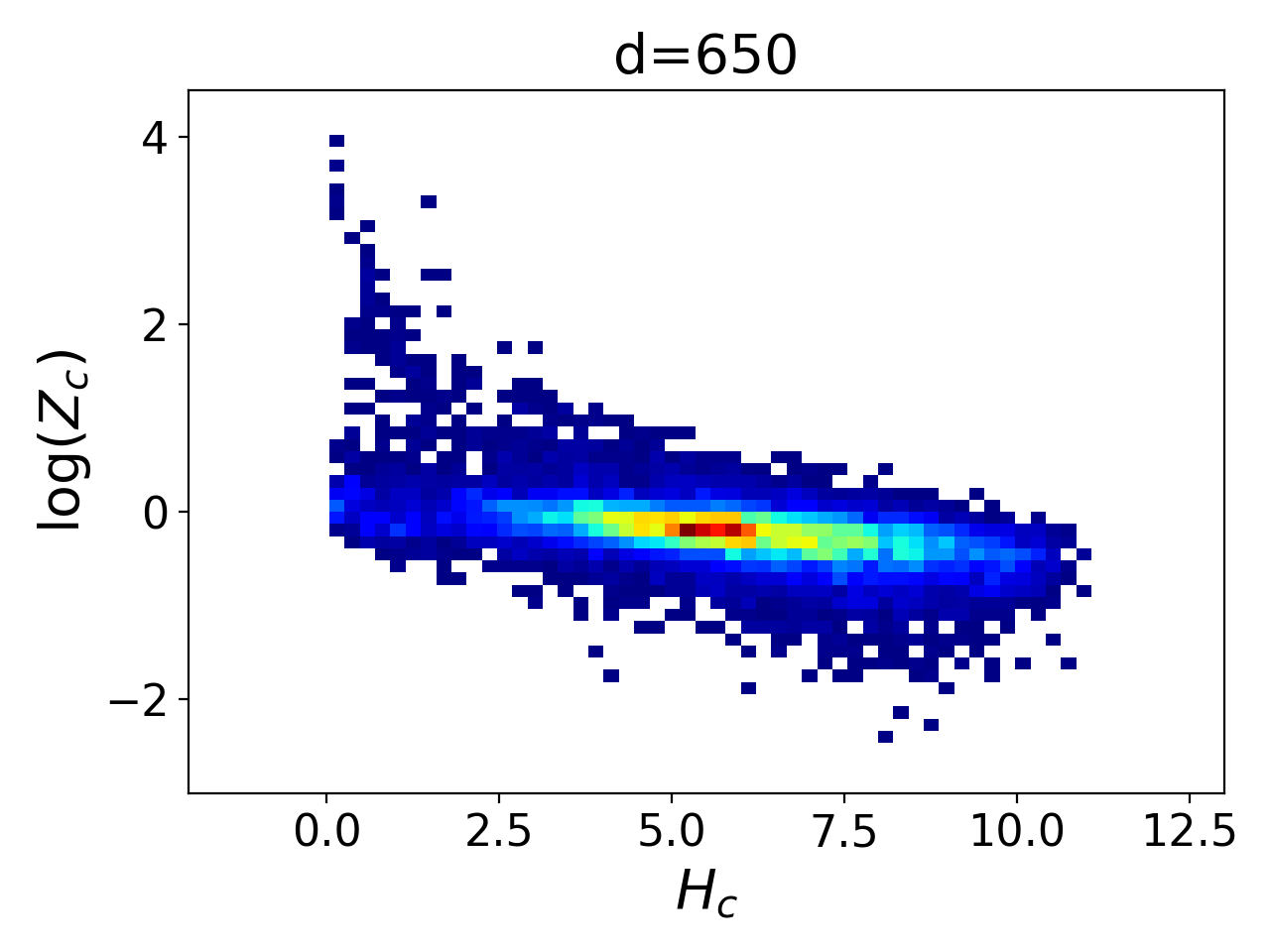} &
\hspace{0.0cm}
\includegraphics[width=5.0cm]{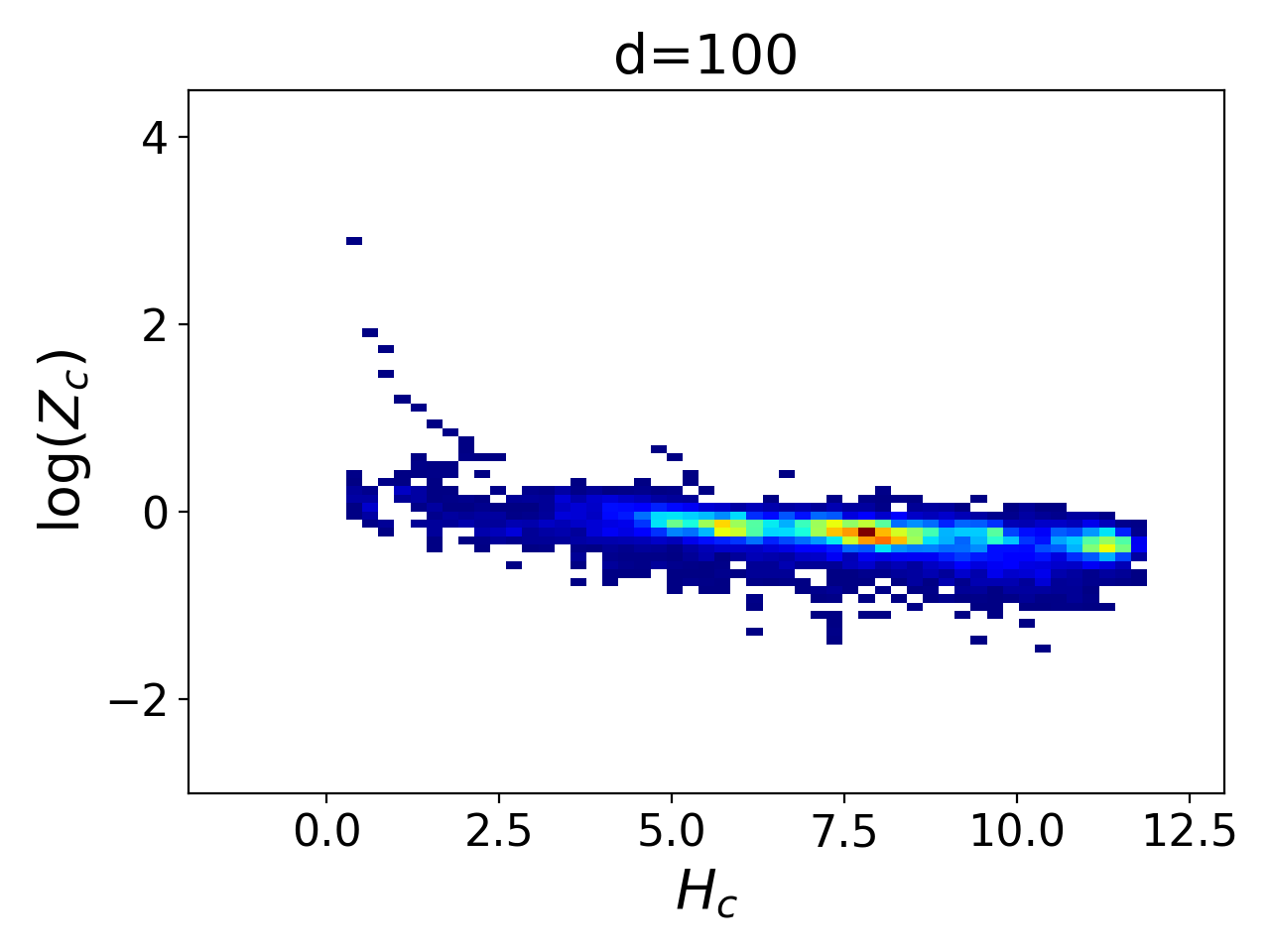} &
\hspace{0.0cm}
\includegraphics[width=5.0cm]{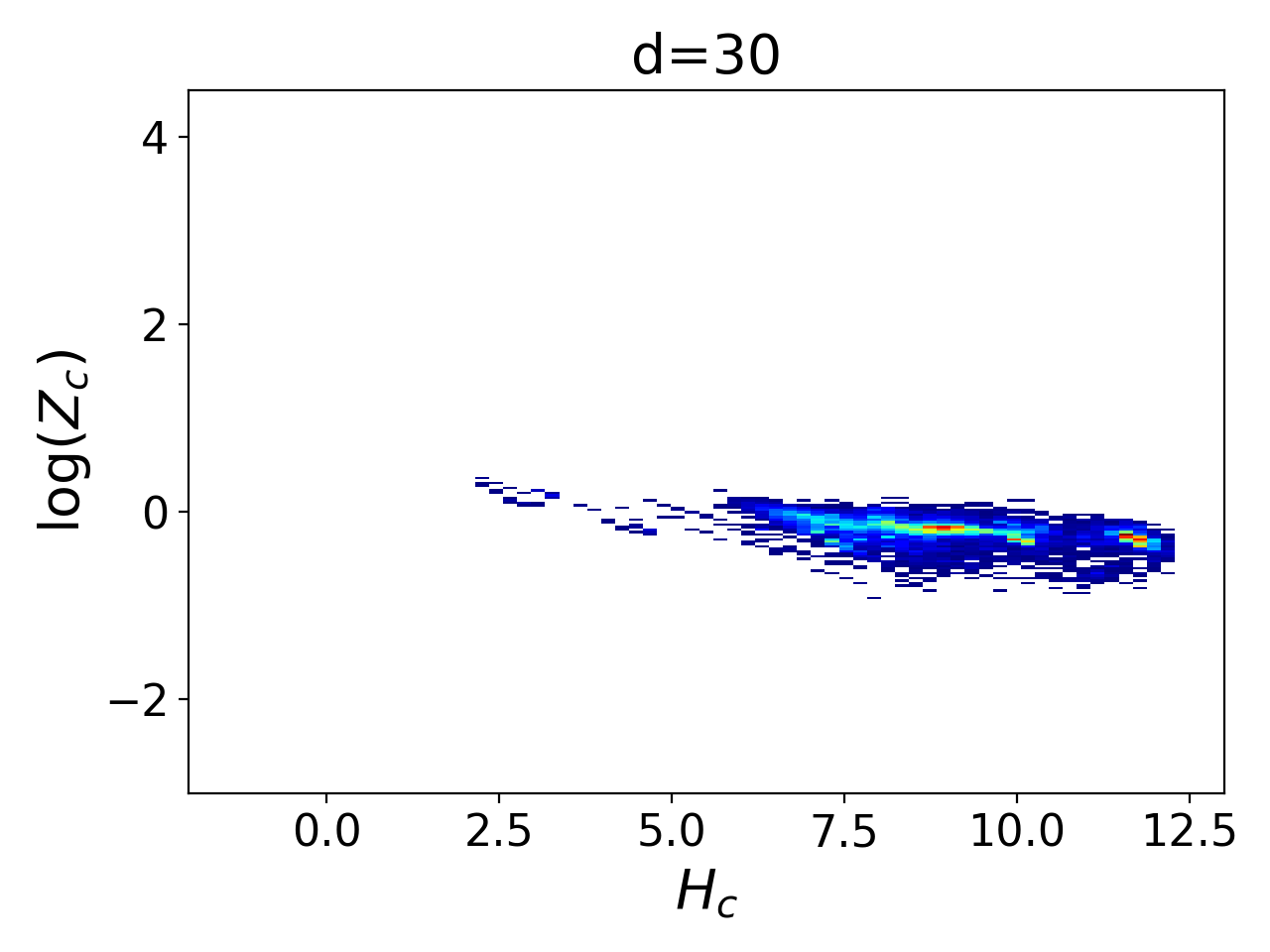}
\end{array}
$
\caption{A 2-dimensional histogram of the normalization term of a predicted distribution as a function of its entropy, as measured over a sample from NCE-LM predictions on the WIKI validation set. Brighter colors denote denser areas.
}
\label{fig:wiki_analysis}
\end{figure*}

\subsection{Analysis}
\label{sec:analysis}
The entropy of the distributions predicted by a language model is a measure of how uncertain it is regarding the identity of the predicted word. Low-entropy distributions would be concentrated around few possible words, while high-entropy ones would be much more spread out. To more carefully analyze the self-normalization properties of NCE-LM and DEV-LM, we computed the Pearson's correlation between the entropy of a predicted distribution $H_c =-\sum_v p(v|c)\log p(v|c)$ and its normalization term, $\log(Z_c)$. As can be seen in Table \ref{tab:entropy_correlation}, it appears that a regularity exists, where the value of $\log(Z_c)$ is negatively correlated with entropy. Furthermore, it seems that, to an extent, the correlation is stronger for larger models.
To further illustrate this regularity, Figure~\ref{fig:wiki_analysis} shows a 2-dimensional histogram of a sample of distributions predicted by NCE-LM . We can see there that particularly low entropy distributions can be associated with very high values of $\log(Z_c)$, deviating a lot from the self-normalization objective of $\log(Z_c)=0$.
Examples for contexts with such low-entropy distributions are: ``During the American Civil [War]'' and ``The United [States]'', where the actual word following the preceding context appears in parenthesis.
This phenomenon is less evident for smaller models, which tend to produce fewer low entropy predictions.

We hypothesize that the above observations could be a contributing factor to our earlier finding that larger models have larger variance in their normalization terms, though it seems to account only for some of that at best. Furthermore, we hope that this regularity could be exploited to improve self-normalization algorithms in the future.

\section{Sentence Completion Challenge}
\label{sec:mscc}

In Section \ref{sec:intrinsic}, we've seen that there may be some trade-offs between perplexity, self-normalization and run-time complexity of language models. While the language modeling method should ultimately to be optimized for each downstream task individually, we follow \newcite{Mnih2012} and use the Microsoft Sentence Completion Challenge \cite{zweig2011microsoft} as an example use case.

The Microsoft Sentence Completion Challenge (MSCC) \cite{zweig2011microsoft} includes 1,040 items. Each item is a sentence with one word replaced by a gap, and the challenge is to identify the word, out of five choices, that is most meaningful and coherent as the gap-filler. 
The MSCC includes a learning corpus of approximately 50 million words.
To use this corpus for training our language models, we split it into sentences, shuffled the sentence order and considered all words with frequency less than 10 as unknown, yielding a vocabulary of about 50K word types.
We used the same settings described in Section \ref{subsec:experimental_settings} to train the language models except that due to the larger size of the data, we ran fewer training iterations.
\footnote{We started with a learning rate of 1 and reduced it by a factor of 2 after each iteration beginning with the very first one.}
Finally, as the gap-filler, we choose the word that maximizes the score of the entire sentence, where a sentence score is the sum of its words' scores. For a normalized language model this score can be interpreted as the estimated log-likelihood of the sentence. 

\begin{table}[t]
\centering
\begin{tabular}{| l || c | c | c | c || c | c | c | c |}
\hline
  & \multicolumn{4}{| c ||}{2 training iterations} &  \multicolumn{4}{| c |}{5 training iterations } \\
\hline
   \rowcolor[gray]{.9}
    & acc-n & $\Delta$acc & perp & $\sigma_z$ &  acc-n & $\Delta$acc & perp & $\sigma_z$\\
\hline

DEV-LM  & 47.6 & +0.4 & 75.3 & 0.10 & 51.9 & -0.7 & 67.5 & 0.10 \\
NCE-R-LM & 46.3 & -0.5 & 78.3 & 0.11 & 51.0 & -0.2 & 70.2 & 0.10\\
NCE-LM & 47.6 & -0.4 & 75.3 & 0.17 & 51.6 & +1.1 & 67.1 & 0.14 \\
SM-LM & 47.0 & \textbf{-2.0} & 73.4 & 1.15 & 51.0 & \textbf{-2.3} & 66.3 & 1.19 \\

 \hline
\end{tabular}
\caption{Microsoft Sentence Completion Challenge (MSCC) results for models with $d=650$ that were trained with 2 and 5 iterations. 'acc-n' denotes the accuracy measure obtained when language model scores are precisely normalized. '$\Delta$acc' denotes the delta in accuracy when unnormalized scores are used instead. 'perp' and $\sigma_z$ denote the mean perplexity and standard deviation of $log(Z_c)$ recorded for the 1,040 answer sentences.}

\label{tab:mscc_results}
\end{table}

The results of the MSCC experiment appear in Table \ref{tab:mscc_results}. Accuracy is the standard evaluation metric for this benchmark (simply the proportion of questions answered correctly). We report this metric when performing the costly test-time score normalization and then the delta observed when using unnormalized scores instead. First, we note that given the same number of training iterations, all methods achieved fairly similar normalized-scores accuracies, as well as perplexity values. At the same time, we do see a notable improvement in both accuracies and perplexities when more training iterations are performed.
Next, with the exception of SM-LM, all of the compared models exhibit good self-normalization properties, as is evident from the low $\sigma_z$ values. There does not seem to be a meaningful accuracy performance hit when using unnormalized-scores for these models, suggesting that this level of self-normalization is adequate for the MSCC task. Finally, as expected, SM-LM exhibits worse self-normalization properties. However, somewhat surprisingly, even in this case, we see a relatively small (though more noticeable) hit in accuracy. This suggests that in some use cases, the level of the language model's self-normalization may have a relatively low impact on the performance of a down-stream task.

\section{Conclusions}

We reviewed and analyzed the two alternative approaches for self-normalization of language models, namely, using Noise Contrastive Estimation (NCE) that is inherently self-normalized, versus adding explicit self-normalizing regularization to a softmax objective function.
Our empirical investigation compared these approaches, and by extending NCE language modeling with a light-weight explicit self-normalization, we also introduced a hybrid model that achieved both good self-normalization and perplexity performance, as well as little dependence of train-time on the size of the vocabulary. 
To put our intrinsic evaluation results in perspective, we used the Sentence Completion Challenge as an example use-case. The results suggest that it would be wise to test the sensitivity of the downstream task to self-normalization, in order to choose the most appropriate method.
Finally, further analysis revealed unexpected correlations between self-normalization and perplexity performance, as well as between the partition function of self-normalized predictions and the entropy of the respective distribution. We hope that these insights would be useful for improving self-normalizing models in future work.

%\bibliography{naaclhlt2018}
\bibliography{paper}
\bibliographystyle{acl}

%\appendix{Appendix}

\end{document}